\documentclass[11pt]{article}

\usepackage[preprint]{acl}

\usepackage{times}
\usepackage{latexsym}

\usepackage[T1]{fontenc}

\usepackage{microtype}

\usepackage{inconsolata}

\usepackage{graphicx}
\usepackage{amsmath,amssymb}

\usepackage{amsthm}
\usepackage[linesnumbered,ruled,vlined]{algorithm2e}
\usepackage{booktabs}
\usepackage{multirow}
\usepackage{tcolorbox}
\usepackage{enumitem}

\usepackage{xcolor}
\usepackage{colortbl}

\definecolor{xred}{HTML}{BD4242}
\definecolor{xblue}{HTML}{C7A085}
\definecolor{xblues}{HTML}{52B256}
\definecolor{xgreen}{HTML}{52B256}
\definecolor{xpurple}{HTML}{7F52B2}
\definecolor{xorange}{HTML}{FD9337}
\definecolor{xdotted}{HTML}{999999}
\definecolor{xgray}{HTML}{777777}
\definecolor{xcyan}{HTML}{80F5DC}
\definecolor{xpink}{HTML}{f690ea}
\definecolor{xgraycyan}{HTML}{82bceb}
\usepackage{listings}
\usepackage{xcolor}

\definecolor{codegreen}{rgb}{0,0.6,0}
\definecolor{codegray}{rgb}{0.5,0.5,0.5}
\definecolor{codepurple}{rgb}{0.58,0,0.82}
\definecolor{backcolour}{rgb}{0.96,0.96,0.96} 

\lstdefinestyle{pythonstyle}{
    backgroundcolor=\color{backcolour},   
    commentstyle=\color{codegreen},
    keywordstyle=\color{magenta},
    numberstyle=\tiny\color{codegray},
    stringstyle=\color{codepurple},
    basicstyle=\ttfamily\footnotesize, 
    breakatwhitespace=false,         
    breaklines=true,                 
    captionpos=b,                    
    keepspaces=true,                 
    numbers=left,                    
    numbersep=5pt,                  
    showspaces=false,                
    showstringspaces=false,
    showtabs=false,                  
    tabsize=4,
    frame=single,                    
    rulecolor=\color{gray!30}        
}

\newcommand\ntfootnote[1]{%
  \begin{NoHyper}
  \renewcommand\thefootnote{}\footnotetext{#1}%
  \addtocounter{footnote}{0}%
  \end{NoHyper}
}

\title{DataChef: Cooking Up Optimal Data Recipes for LLM Adaptation \\via Reinforcement Learning}

\usepackage{fontawesome5}

\author{\textbf{Yicheng Chen$^{1,2}$, Zerun Ma$^{2}$, Xinchen Xie$^{2}$, Yining Li$^{2\dag}$, Kai Chen$^{2\dag}$} \\
{$^{1}$Fudan University}
{$^{2}$Shanghai AI Laboratory} \\
\faGithub\ \textbf{Github}: \href{https://github.com/yichengchen24/DataChef}{https://github.com/yichengchen24/DataChef}
} 

\begin{document}
\twocolumn[{%
  \renewcommand\twocolumn[1][]{#1}%
  \maketitle
    \vspace{-40pt}
    \captionsetup{type=figure}
    \centering
    \includegraphics[width=.85\textwidth]{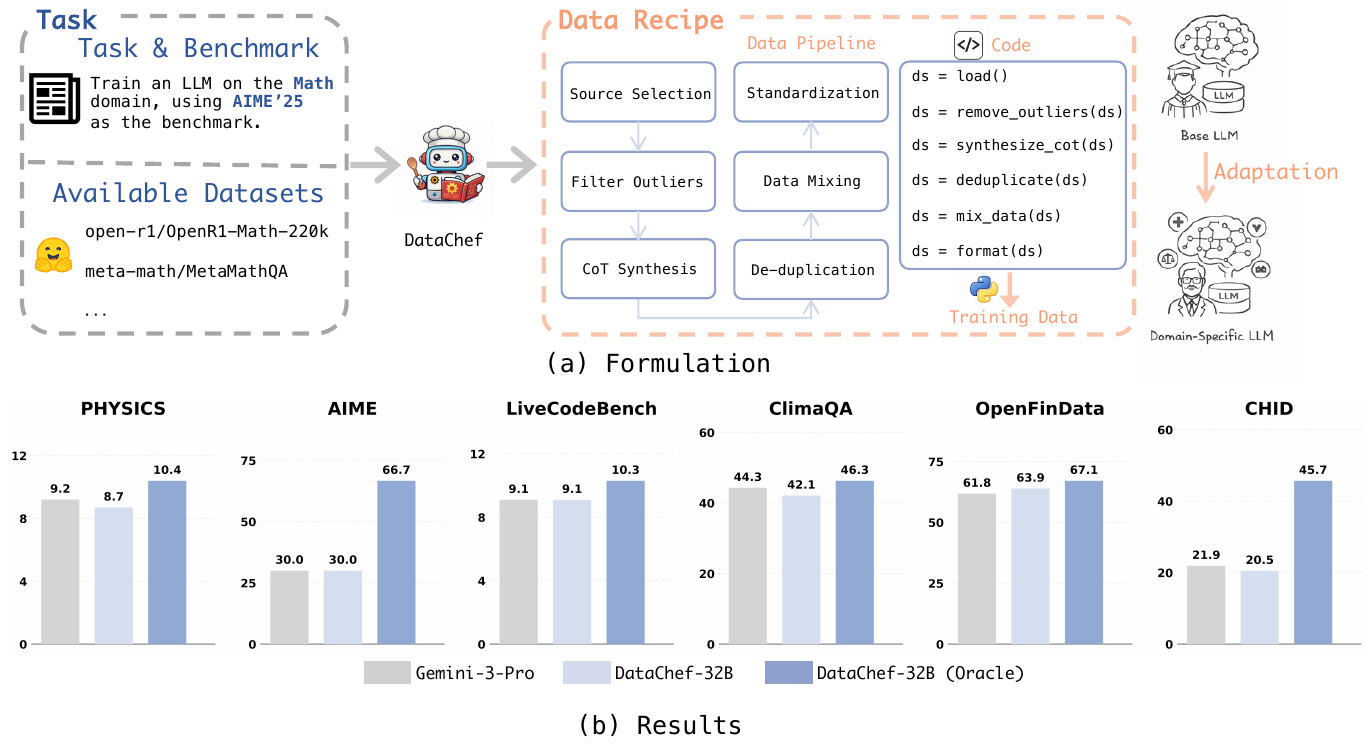}
    \vspace{-8pt}
    \caption{(a) \textbf{Formulation}. Given a task instruction, evaluation protocol, and raw data sources, a model is required to generate a data recipe, including an executable pipeline and the resulting training dataset, for LLM adaptation. (b) \textbf{Main results}. DataChef matches the performance of recipes from Gemini-3-Pro across six held-out tasks. See details in Sec.~\ref{sec:main-results}.}
    \label{fig:teaser}
    \vspace{12pt}
}]
\ntfootnote{$^{\dag}$ Corresponding Author.}
\begin{abstract}
In the current landscape of Large Language Models (LLMs), the curation of large-scale, high-quality training data is a primary driver of model performance.
A key lever is the \emph{data recipe}, which comprises a data processing pipeline to transform raw sources into training corpora.
Despite the growing use of LLMs to automate individual data processing steps, such as data synthesis and filtering, the overall design of data recipes remains largely manual and labor-intensive, requiring substantial human expertise and iteration.
To bridge this gap, we formulate \emph{end-to-end data recipe generation} for LLM adaptation. Given a target benchmark and a pool of available data sources, a model is required to output a complete data recipe that adapts a base LLM to the target task.
We present DataChef-32B, which performs online reinforcement learning using a proxy reward that predicts downstream performance for candidate recipes.
Across six held-out tasks, DataChef-32B produces recipes that yield performance comparable to those curated by human experts.
Notably, the recipe from DataChef-32B adapts Qwen3-1.7B-Base to the math domain, achieving 66.7 on AIME'25 and surpassing the official post-training checkpoint (Qwen3-1.7B).
This work sheds new light on automating LLM training and developing self-evolving AI systems.
\end{abstract}
    
\section{Introduction}
The rapid evolution of LLMs~\cite{deepseekv32,gpt-5} has precipitated a shift toward data-centric AI~\cite{DCAI}, identifying the composition and quality of training data as decisive factors in shaping model performance.
In practice, constructing effective training data requires a well-designed multi-stage pipeline that processes heterogeneous raw data through a sequence of operations, such as transformation, filtering, mixing, synthesis, and refinement, tailored to specific training goals or stages~\cite{qwen3,opendataarena}.
In this work, we formalize such processing pipelines with the resulting dataset as a \emph{data recipe}.

Currently, formulating an effective data recipe relies heavily on human heuristics, where experts manually orchestrate data processing operations and iteratively refine them based on empirical feedback~\cite{penedo2025fineweb2,gururajan2024aloe}. 
While LLMs are widely used to automate individual pipeline components, such as data filtering~\cite{Deita,zhang-etal-2025-autonomous} and synthesis~\cite{mitra2024agentinstructgenerativeteachingagentic,huang2024mustardmasteringuniformsynthesis}, they typically operate under rigid, hand-crafted prompts or algorithms.
Recent studies have explored automating data pipeline orchestration to reduce the reliance on manual effort.
In particular, Data-Juicer Sandbox~\cite{DataJuicerSandbox} proposes a Probe-Analyze-Refine workflow to identify the most impactful operators from a predefined operation pool, combine effective operations, and optimize data utilization through systematic experiments in data processing, model training, and evaluation with model performance as feedback.
However, the continuous scaling of data and model sizes, coupled with the increasing complexity of processing operations, renders an exhaustive exploration of the combinatorial space of data recipes infeasible.
Therefore, an essential question arises: can AI systems automatically generate a data recipe for training LLMs, including the orchestration of data pipelines and the implementation of each operation, in a cost-efficient way?

To bridge this gap, we introduce a new task: \emph{end-to-end data recipe generation} for LLM adaptation.
As shown in Fig~\ref{fig:teaser}(a), given a target benchmark and a pool of available data sources, the objective is to generate a data recipe by specifying the precise data processing pipeline to yield training data for adapting an LLM to the target task.
This task requires sophisticated reasoning to analyze heterogeneous data sources, apply domain-specific processing operations, and generate executable code. 
While reinforcement learning with verifiable rewards (RLVR) has proven effective in enhancing LLM reasoning in domains such as coding~\cite{AceCoder} and mathematics~\cite {yeo2025demystifyinglongchainofthoughtreasoning}, applying this paradigm to our task poses two key challenges:
(1) \textbf{Data absence}: As a novel task, there are no curated datasets or standardized evaluation benchmarks for data recipe generation.
(2) \textbf{Expensive and delayed supervision}: While downstream performance naturally serves as the reward signal, it is impractical to incorporate full LLM training into an online RL loop.

To address these challenges, we curate a comprehensive task pool comprising 31 widely used benchmarks across 19 distinct domains. These domains encompass reasoning-heavy fields, such as mathematics and coding, as well as knowledge-centric fields, such as finance, medicine, and natural sciences. The pool is partitioned into 25 training tasks and 6 held-out evaluation tasks, with each task supported by 8--15 source training datasets.
We further propose a \emph{Data Verifier} that estimates the utility of processed data without requiring model training, providing a low-latency proxy reward for scalable online RL.
Empirical validation demonstrates that our Data Verifier correlates well with downstream performance and exhibits superior robustness across diverse tasks compared to existing data scoring metrics~\cite{IFD,VendiScore}.
Leveraging this task pool and proxy reward, we present DataChef-32B, an LLM specialized in generating optimal data recipes.

Extensive evaluations on 6 held-out tasks demonstrate that DataChef-32B matches the capabilities of the state-of-the-art proprietary model, Gemini-3-Pro, as illustrated in Fig.~\ref{fig:teaser}(b).
Furthermore, recipes generated by DataChef-32B outperform the SOTA selection algorithm, DEITA~\cite{Deita}, on most tasks, highlighting the superior potential of exploring a vast coding space over relying on hand-designed selection heuristics.
Notably, our recipes adapt Qwen3-1.7B-Base to achieve 66.7 on AIME'25~\cite{aime} and 46.3 on ClimaQA~\cite{climaqa}, respectively, surpassing the official Qwen3-1.7B checkpoint with industry-level post-training on expert-curated data recipes.

In summary, our contributions are as follows:
\begin{enumerate}[label={\bf {{$\bullet$}}}, leftmargin=*, topsep=0.5ex, itemsep=-0.5ex, partopsep=0.75ex, parsep=0.75ex, partopsep=0pt, wide, labelindent=0pt]
    \item We formulate a new task, end-to-end data recipe generation for LLM adaptation, requiring models to automatically generate data recipes from a benchmark and available data sources.
    \item We construct a large-scale and diverse data pool covering 19 domains, 31 benchmarks, and 257 datasets to facilitate research in this area.
    \item We propose an efficient learning framework with a proxy reward that enables scalable online RL. Experiments show that DataChef-32B achieves performance comparable to top-tier proprietary models on the data recipe generation task.
\end{enumerate}
\section{Related Work}

\noindent 
\textbf{Data Pipelines.}
%
Many existing approaches rely on human experts to design individual data processing heuristics, including data mixing~\cite{regmix}, data sampling~\cite{xu2023demystifying,MIG}, and data synthesis~\cite{ACP}.
General-purpose data processing frameworks~\cite{DataJuicer,park2025dataverse} provide standardized modules and scalable pipeline construction for large-scale data processing, and are adopted to curate large-scale, high-quality training data, such as FinWeb2~\cite{penedo2025fineweb2} for multilingual pre-training and Aloe~\cite{gururajan2024aloe} for medical-domain fine-tuning.
However, their efficiency remains constrained by the manual pipeline design and iterative trial-and-error on downstream tasks.
Data-Juicer Sandbox~\cite{DataJuicerSandbox} marking a step further towards automated data pipeline construction by employing a Probe-Analyze-Refine workflow to assess operator effectiveness, but still relies on feedback derived from downstream model training, which is time and computation-consuming.
%
%
%
In contrast, our work aims to end-to-end generate data recipes from scratch.

\noindent
\textbf{LLM Agents for Data Science.}
LLM-based agent systems have emerged as powerful tools for automating data science workflows, including data analysis, modeling, and visualization.
Most existing approaches~\cite{hollmann2023largelanguagemodelsautomated,AutoKaggle,DataInterpreter} rely on prompt-based approaches, where complex tasks are decomposed and solved according to heuristically designed workflows.
AIDE~\cite{aide} and SELA~\cite{SELA} further adopt iterative exploration and refinement through trial-and-error execution.
Yet such prompt-driven strategies remain largely static and are constrained by the inherent knowledge limitations of LLMs.
To alleviate these limitations, some studies incorporate external knowledge via search-based methods, leveraging offline repositories such as Kaggle solutions and research papers~\cite{DSAgent,automind,KompeteAI} or online web search~\cite{MLESTAR}.
Another line of work~\cite{MLAgent,DeepAnalyze} explores learning-based agents, where agents improve performance through interaction and experience.
However, these methods are typically evaluated on well-defined Kaggle competitions~\cite{MLEBench,datascibench,DSBench} with static datasets, and even with curated initial code. 
In this work, we address an open-ended setting, taking arbitrary tasks and available datasets as input and directly generating data recipes for LLM training.

\noindent
\textbf{Data Evaluation.}
Training and evaluating LLMs require significantly more computational resources, motivating the use of lightweight proxies to assess model performance~\cite{DataJuicerSandbox}.
Existing data evaluation approaches~\cite{qin2024unleashing,zhang2025survey} can be broadly categorized into three groups. (1) Indicator-based methods~\cite{IFD,VendiScore} define handcrafted metrics to quantify properties such as diversity, complexity, and relevance. 
(2) Model-based methods~\cite{CaR,Deita} train predictive models to estimate data quality. 
(3) LLM-as-a-Judge approaches~\cite{alpagasus} prompts powerful LLMs to evaluate data according to specific protocols.
However, the correlation between data assessment scores and downstream model performance remains underexplored.
Prior work typically validates evaluators by comparing specific data selections against baselines, rather than through systematic correlation analysis.
To bridge this gap, we conduct a comprehensive study of representative assessment methods, evaluating their alignment with model performance across diverse fine-tuning tasks.

%

\section{Methodology}

In this section, we first formalize some core concepts and define the data recipe generation task in Sec~\ref{sec:problem-formulation}. Then, we introduce the specific data pool constructed for this study in Sec~\ref{sec:task-construction}. Finally, we present our learning framework in Sec~\ref{sec:framework}.

\begin{figure*}[t]
    \centering
    \includegraphics[width=0.8\linewidth]{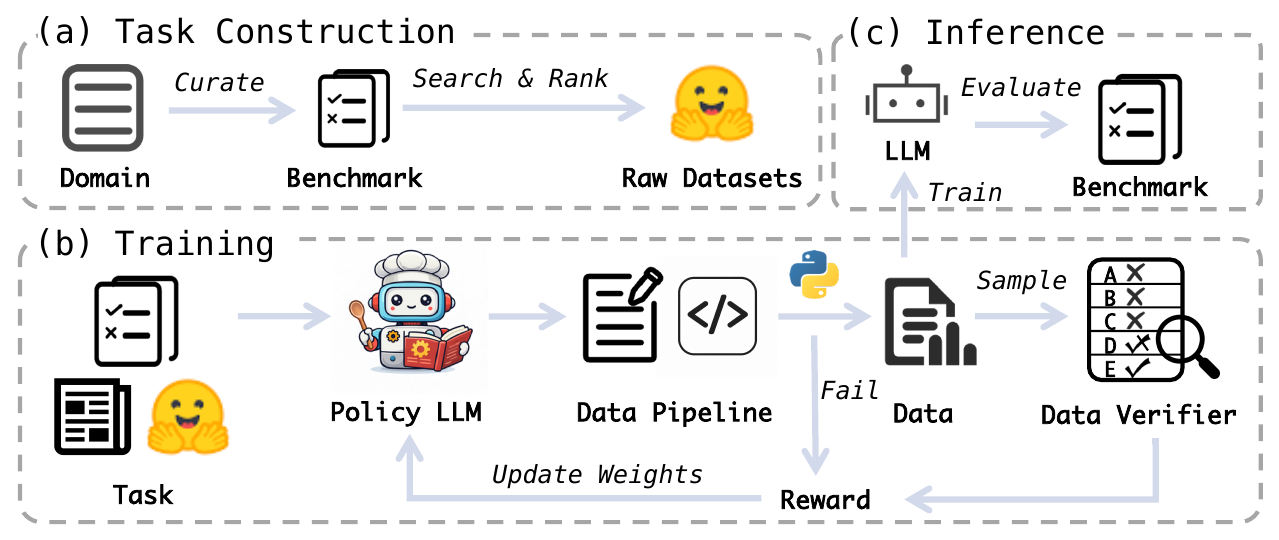}
    \vspace{-4pt}
    \caption{\textbf{Illustration of task construction and DataChef framework}. Given a task, a policy LLM generates a data recipe, which is executed to produce a training dataset. The Data Verifier then evaluates a sampled subset to provide a scalar reward, guiding the policy update via GRPO to optimize for data quality and executability.}
    \label{fig:framework}
    \vspace{-12pt}
\end{figure*}

\subsection{Problem Formulation}
\label{sec:problem-formulation}
The goal of our method is to automatically generate a data recipe given a specific task. We formulate a \textbf{task} as a triplet $T = (I, \tau, \mathcal{D})$, where $I$ is a natural language instruction, including description of the task requirement, along with meta-information of data sources and evaluation protocol, $\mathcal{D}$ denotes the set of available raw data sources, and $\tau$ is an evaluation metric that maps any model $\mathcal{M}$ to a scalar performance score $\tau(\mathcal{M})\in\mathbb{R}$. A \textbf{data recipe} is formulated as $r = (g,d)$, where $g \in \mathcal{G}$ is a data pipeline and $d=g(\mathcal{D})$ is the resulting training dataset. In our experiments, the data pipeline is implemented as Python scripts.

Let $\mathcal{M}_\theta$ denote a language model. We use $\theta_{d}$ to present the parameters fine-tuned on a dataset $d$.
We aim to learn a policy $\pi_{\phi}(r\mid T)$ that generates data recipes to maximize the expected downstream performance of the trained model. Formally, the objective function is defined as:
\begin{equation}
\label{eq:recipe-obj}
\mathcal{J}(\phi) = \mathbb{E}_{r\sim\pi_\phi(\cdot\mid T)}[\tau(\mathrm{LM}_{\theta_d})]
\end{equation}

\subsection{Task Pool Construction}
\label{sec:task-construction}
\noindent
\textbf{Seed Task Curation.} 
We construct a diverse task pool encompassing 19 heterogeneous domains, including reasoning, coding, and knowledge-intensive fields such as healthcare, finance, and natural science.
As shown in Fig~\ref{fig:framework}(a), for each domain, we curate representative benchmarks (e.g., GSM8K and AIME'25 for mathematics), totaling 31 benchmarks.
For each benchmark, we retrieve relevant candidate datasets from Hugging Face, prioritizing those with high community engagement (downloads and likes), yielding a repository of 257 distinct data sources.
From this collection, we construct 25 seed tasks for training and reserve 6 held-out tasks (3 in-domain, 3 out-of-domain) for evaluation.
Details of the selected benchmarks and automatic retrieval procedure for candidate datasets are provided in Appx.~\ref{appx:task-pool}.

\noindent
\textbf{Training Task Augmentation.}
To facilitate robust policy learning, we expand the 25 seed tasks into a large-scale training set $\mathcal{T}_{\mathrm{train}}$.
We employ a probabilistic sampling strategy where a benchmark $\tau$ is selected proportional to its source count $|\mathcal{D}|$, followed by uniform sampling of a subset $\mathcal{D}' \subseteq \mathcal{D}$ to form a new instance $T'=(I', \tau, \mathcal{D}')$. After deduplication, the expansion strategy yields 5K unique task instances.

\subsection{End-to-end Data Recipe Generation}
\label{sec:framework}
\noindent
\textbf{Framework Overview.}
As illustrated in Fig.~\ref{fig:framework}, our framework optimizes the policy $\pi_{\phi}$ to generate high-quality data recipes.
Given a task $T$, the policy generates a data pipeline $g$, which consists of a natural language plan for orchestrating data pipelines and its corresponding implementation as an executable code block.
During training, the pipeline transforms raw data sources $\mathcal{D}$ into a training dataset $d$, which is then evaluated by the Data Verifier to guide policy updates via reinforcement learning.
During inference, the data recipe is directly used for downstream model adaptation.

\noindent
\textbf{Cold-start Initialization.}
Training the policy from scratch using RL is non-trivial due to the low executability of data recipes, leading to sparse, high-variance rewards and ineffective exploration~\cite{deepseek-math,MLAgent}.
To mitigate this, we employ a cold-start Supervised Fine-Tuning (SFT) phase.
We observe that decoupling reasoning and coding yields superior inference-time performance (detailed in Sec.~\ref{sec:coll}). 
Therefore, we construct a demonstration set using a decoupled generation process: a strong reasoning model proposes plans, and a specialized coding model implements them.
Rejection sampling is used to retain only high-quality rollouts based on execution success and data quality.
Initializing $\pi_{\phi}$ on this curated dataset equips the policy with a foundational capability for code generation, significantly stabilizing the subsequent RL phase.

\definecolor{MyGray}{HTML}{F9EBE3}
\definecolor{HeaderGray}{gray}{0.96}

\begin{table*}[!t]
\centering
\caption{\textbf{Main Results on six held-out tasks}. We report the mean Data Verifier Score $\mathrm{DVS}_{\mathrm{avg}@32}$ and the Downstream Benchmark Score $\mathrm{DBS}$, where the Average column presents $\mathrm{DBS}_\mathrm{norm}$ as a normalized score relative to \textsc{Single-Source}$_{\mathrm{best}}$ (100.0). Qwen3-Next-80B $\oplus$ Kimi-K2 denotes a combination using Qwen3-Next-80B for reasoning and Kimi-K2-Instruct for coding. DataChef-32B achieves performance comparable to the closed-source Gemini-3-Pro and significantly outperforms other open-source baselines across all settings.}
\vspace{-4pt}
\label{tab:main-results}
\renewcommand{\arraystretch}{1.2}
\setlength{\tabcolsep}{5pt}

\scalebox{0.66}{
\begin{tabular}{l cc cc cc cc}
\toprule[1.5pt]

\multicolumn{9}{c}{\small \textbf{\textit{Note: All Methods are fine-tuned from Qwen3-1.7B-Base}}} \\
\midrule[1pt]

\rowcolor{gray!10} \multicolumn{9}{l}{\textbf{\textit{Indomain Tasks}}} \\ 

\multirow{2}{*}{\textbf{Method}} & 
\multicolumn{2}{c}{\textbf{\textsc{PHYSICS}}} &  
\multicolumn{2}{c}{\textbf{\textsc{AIME}}} &  
\multicolumn{2}{c}{\textbf{\textsc{LiveCode}}} &  
\multicolumn{2}{c}{\textbf{Average}} \\
\cmidrule(lr){2-3} \cmidrule(lr){4-5} \cmidrule(lr){6-7} \cmidrule(lr){8-9}

 & \scriptsize $\mathrm{DVS}_{\mathrm{avg}@32} \uparrow$ & \scriptsize $\mathrm{DBS} \uparrow$ & \scriptsize $\mathrm{DVS}_{\mathrm{avg}@32} \uparrow$ & \scriptsize $\mathrm{DBS} \uparrow$ & \scriptsize $\mathrm{DVS}_{\mathrm{avg}@32} \uparrow$ & \scriptsize $\mathrm{DBS} \uparrow$ & \scriptsize $\mathrm{DVS}_{\mathrm{avg}@32} \uparrow$ & \scriptsize $\mathrm{DBS}_{\mathrm{norm}} \uparrow$ \\
\midrule[1pt]

\rowcolor{gray!10} \multicolumn{9}{l}{\textit{\textbf{Model Reference}}} \\
Qwen3-1.7B-Base  & - & 0.8 & - & 20.0 & - & 1.7  & - & 25.5  \\
Qwen3-1.7B       & - & 20.8& - & 33.3 & - & 25.7 & - & 123.3 \\
\midrule

\rowcolor{gray!10}  \multicolumn{9}{l}{\textit{\textbf{Human Preprocess $\oplus$ Data Selection}}} \\
\textsc{Single-Source}$_{\mathrm{avg}}$    & - & 6.1 & - & 23.4 & - & 6.3  & - & 63.9  \\
\textsc{Single-Source}$_{\mathrm{best}}$   & - & 8.5 & - & 39.6 & - & 10.3 & - & 100.0 \\
\textsc{Algorithm}$_{\mathrm{IFD}}$   & - & 7.0 & - & 3.3 & - & 4.6 & - & 45.3 \\
\textsc{Algorithm}$_{\mathrm{DEITA}}$ & - & 7.5 & - & 6.7 & - & 10.9& - & 70.5 \\
\midrule

\rowcolor{gray!10} \multicolumn{9}{l}{\textit{\textbf{LLM-as-a-Chef}}} \\
Qwen3-32B     & 11.0 & 5.9 & 31.5 & 13.3 & 24.3 & 8.0 & 22.3 & 56.7 \\
Kimi-K2       & 19.7 & 9.0 & 35.4 & 20.0 & 19.3 & 9.7 & 24.8 & 83.7 \\
Qwen3-Next-80B $\oplus$ Kimi-K2& 48.7 & 8.9 & 78.3 & 23.3 & 39.2 & 7.4 & 55.4 & 78.6 \\
Gemini-3-Pro  & 69.7 & 9.2 & 80.7 & 30.0 & 53.6 & 9.1 & 68.0 & 91.2 \\

\rowcolor{MyGray} \textbf{DataChef-32B}          & 61.4 & 8.7 & 84.7 & 30.0 & 45.8 & 9.1 & 64.0 & 89.3 \\
\rowcolor{MyGray} \textbf{DataChef-32B (Oracle)} & -    & 10.4& -    & 66.7 & -    & 10.3& -    & 130.3\\

\midrule[2pt] 

\rowcolor{gray!10} \multicolumn{9}{l}{\textbf{\textit{Out-of-domain Tasks}}} \\

\multirow{2}{*}{\textbf{Method}} & 
\multicolumn{2}{c}{\textbf{\textsc{ClimaQA}}} & 
\multicolumn{2}{c}{\textbf{\textsc{OpenFin}}} & 
\multicolumn{2}{c}{\textbf{\textsc{CHID}}} & 
\multicolumn{2}{c}{\textbf{Average}} \\
\cmidrule(lr){2-3} \cmidrule(lr){4-5} \cmidrule(lr){6-7} \cmidrule(lr){8-9}

 & \scriptsize $\mathrm{DVS}_{\mathrm{avg}@32} \uparrow$ & \scriptsize $\mathrm{DBS} \uparrow$ & \scriptsize $\mathrm{DVS}_{\mathrm{avg}@32} \uparrow$ & \scriptsize $\mathrm{DBS} \uparrow$ & \scriptsize $\mathrm{DVS}_{\mathrm{avg}@32} \uparrow$ & \scriptsize $\mathrm{DBS} \uparrow$ & \scriptsize $\mathrm{DVS}_{\mathrm{avg}@32} \uparrow$ & \scriptsize $\mathrm{DBS}_{\mathrm{norm}} \uparrow$ \\
\midrule[1pt]

\rowcolor{gray!10} \multicolumn{9}{l}{\textit{\textbf{Model Reference}}} \\
Qwen3-1.7B-Base & - & 15.1 & - & 33.7 & - & 14.2 & - & 35.9  \\
Qwen3-1.7B      & - & 44.2 & - & 73.4 & - & 59.8 & - & 101.0 \\
\midrule

\rowcolor{gray!10} \multicolumn{9}{l}{\textit{\textbf{Human Preprocess $\oplus$ Data Selection}}} \\

\textsc{Single-Source}$_{\mathrm{avg}}$    & - & 26.0 & - & 41.7 & - & 49.3 & - & 65.1  \\
\textsc{Single-Source}$_{\mathrm{best}}$   & - & 43.6 & - & 63.7 & - & 70.3 & - & 100.0 \\
\textsc{Algorithm}$_{\mathrm{IFD}}$   & - & 37.9 & - & 60.8 & - & 56.6 & - & 87.7 \\
\textsc{Algorithm}$_{\mathrm{DEITA}}$ & - & 38.0 & - & 63.2 & - & 63.7 & - & 92.4 \\

\midrule

\rowcolor{gray!10} \multicolumn{9}{l}{\textit{\textbf{LLM-as-a-Chef}}} \\
Qwen3-32B    & 20.6 & 35.6 & 34.9 & 23.8 & 1.1  & 12.3 & 18.9 & 45.5 \\
Kimi-K2      & 18.3 & 41.8 & 51.5 & 46.5 & 23.1 & 3.9  & 31.0 & 58.2 \\
Qwen3-Next-80B $\oplus$ Kimi-K2& 41.5 & 42.6 & 54.7 & 64.0 & 8.6  & 4.1  & 34.9 & 68.0 \\
Gemini-3-Pro & 58.4 & 44.3 & 54.9 & 61.8 & 29.7 & 21.9 & 47.6 & 76.6 \\

\rowcolor{MyGray} \textbf{DataChef-32B}          & 57.3 & 42.1 & 67.0 & 63.9 & 7.9 & 20.5 & 44.1 & 75.4 \\
\rowcolor{MyGray} \textbf{DataChef-32B (Oracle)} & -    & 46.3 & -    & 67.1 & -   & 45.7 & -    & 92.2 \\
\bottomrule[1.5pt]
\end{tabular}
}
\vspace{-8pt}
\end{table*}

\noindent
\textbf{Reward Modeling.}
Ideally, the reward signal would be the downstream performance $\tau(\mathcal{M}_{\theta_d})$.
However, using this as an online reward is computationally prohibitive due to the cost of repeated model training and evaluation.
Instead, we design a computationally efficient surrogate reward based on the quality of the generated dataset $d$.
Inspired by rubrics-based rewards~\cite{gunjal2025rubricsrewardsreinforcementlearning}, we employ a strong LLM as a \textit{Data Verifier} to classify each instance $x \in d$ into one of five categories with assigned scalar scores $s(x)$:
\begin{enumerate}[label={\bf{{$\bullet$}}}, leftmargin=*, topsep=0.2ex, itemsep=-0.5ex, partopsep=0.75ex, parsep=0.75ex, partopsep=0pt, wide, labelindent=0pt]
    \item \textit{Invalid} ($0$): Samples with missing essential information or severe repetition.
    \item \textit{Format Error} ($0$): Samples violating explicit output format constraints.
    \item \textit{Incorrect} ($0$): Samples containing factual errors or wrong answers.
    \item \textit{Task Mismatch} ($0.4$): Valid samples that are semantically irrelevant to the target task $I$.
    \item \textit{Pass} ($1.0$): High-quality samples that satisfy all criteria.
\end{enumerate}
To ensure computational efficiency during online training, we estimate the dataset quality by randomly sampling a subset $\hat{d} \subset d$.
Let $\bar{s}(\hat{d})$ be the average instance score over this sampled subset. We define the final recipe reward $\mathcal{R}(r)$ by incorporating penalties for execution failures:
\begin{equation}
\mathcal{R}(r) = \begin{cases}
    -\lambda_{\emptyset}, & \text{if}\ d = \emptyset\ \text{\small(execution failure)},\\
    -\lambda_{\mathrm{fmt}}, & \text{if}\ d\ \text{\small{violates training format}}, \\
    \bar{s}(\hat{d}), & \text{\small{otherwise}},
\end{cases}
\label{eq:reward}
\end{equation}
where $\lambda_{\emptyset}$ and $\lambda_{\mathrm{fmt}}$ are positive penalty coefficients. Please refer to Appx.~\ref{appx:prompt} for a detailed description of the category definitions used in the prompt and to Appx.~\ref{appx:comp-resources} for computational cost analysis.

\noindent
\textbf{Reinforcement Learning.}
We employ Group Relative Policy Optimization (GRPO) for policy optimization.
For each task $T\sim\mathcal{T}_{\mathrm{train}}$, we sample a group of $G$ candidate data recipes $\{r_i\}_{i=1}^{G}$ from the current policy $\pi_{\phi_{\mathrm{old}}}$.
The policy parameters are optimized by maximizing the following objective:
\begin{equation}
\small
\label{eq:grpo}
\begin{aligned}
\mathcal{J}(\phi)=
\mathbb{E}\Bigg[
\frac{1}{G}\sum_{i=1}^{G}
\min\Big(
& \rho_i A_i,\;
\mathrm{clip}(\rho_i,1-\epsilon,1+\epsilon)\,A_i
\Big) \\
& -\beta\, D_{\mathrm{KL}}\!\Big(\pi_\phi\,\|\,\pi_{\mathrm{ref}}\Big)
\Bigg]
\end{aligned}
\end{equation}
where $\rho_i=\frac{\pi_\phi(r_i\mid T)}{\pi_{\phi_{\mathrm{old}}}(r_i\mid T)}$ is the importance ratio, $A_i=\frac{\mathcal{R}(r_i)-\mu}{\sigma+\delta}$ is the group-relative advantage, $\epsilon$ is the clipping parameter, $\pi_{\mathrm{ref}}$ is a fixed reference policy, and $\beta$ controls KL regularization.

\section{Experiments}

\subsection{Setups}

\begin{figure*}[t]
    \centering
    \includegraphics[width=0.98\linewidth]{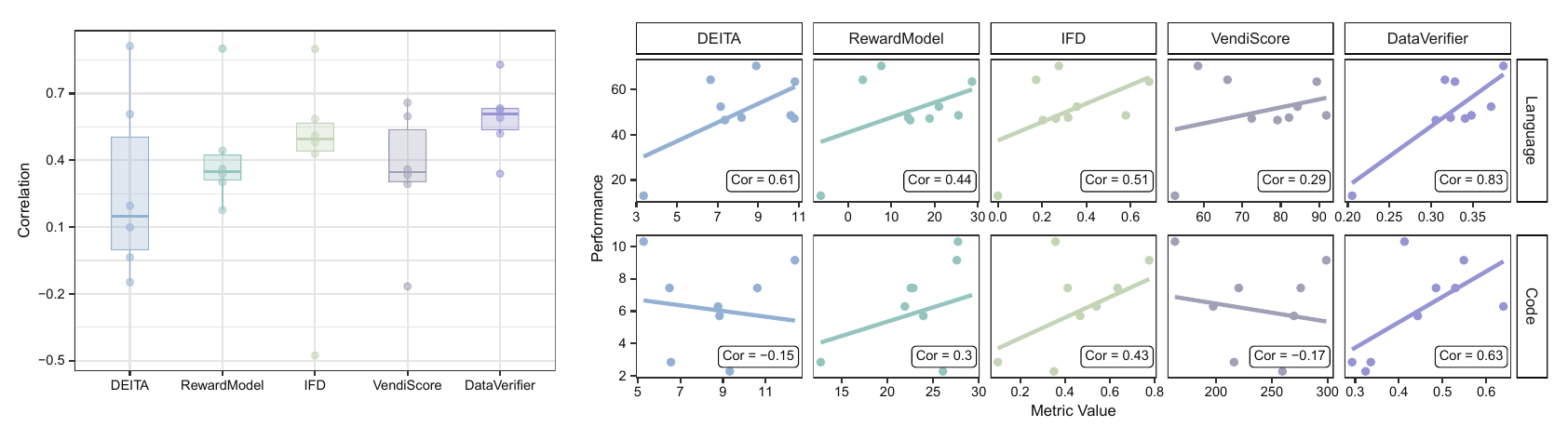}
    \vspace{-8pt}
    \caption{\textbf{Correlation analysis of data evaluation metrics}. (left) We summarize the Pearson correlation coefficients across all six evaluated tasks. (right) We detail the relationship between metric scores (X-axis) and downstream performance (Y-axis) on Language and Code tasks. The Data Verifier maintains a strong, consistent positive correlation across disparate domains. Please refer to Table~\ref{tab:correlation_results} and Fig.~\ref{fig:appx-cor-all} in Appx.~\ref{appx:additional-results-on-correlation-analysis} for complete results.}
    \label{fig:cor-analysis}
    \vspace{-6pt}
\end{figure*}


\noindent
\textbf{Training.}
For cold-start SFT, we train Qwen3-32B~\cite{qwen3} on 5K high-quality synthetic instances for 2 epochs, utilizing a learning rate of 2e-5 and a batch size of 32.
In the RL phase, we further optimize the SFT checkpoint using GRPO~\cite{deepseek-math} for 1 epoch on the same dataset, with a learning rate of 5e-7.
During RL, the rollout batch size is set to $128$ with a temperature of $1.0$, and we sample 8 candidate data recipes per task.

\noindent
\textbf{Evaluation Set.}
We evaluate on 6 held-out tasks: 3 in-domain tasks and 3 out-of-domain tasks. 
Notably, these in-domain evaluation tasks share domains with the training set but remain strictly unseen during training.
The in-domain benchmarks include PHYSICS~\cite{physics}, AIME'25~\cite{aime}, and LiveCodeBench v6~\cite{livecodebench}; the out-of-domain benchmarks are OpenFinData~\cite{openfindata}, ClimaQA~\cite{climaqa}, and CHID~\cite{chid}.

\noindent
\textbf{Metrics.}
Executing recipes and performing downstream fine-tuning and evaluation are compute-intensive, rendering large-scale end-to-end evaluation impractical.
Accordingly, for each evaluation task, we generate a candidate set of $N=32$ independent data recipes.
Based on this candidate set, we report two metrics:
(1) $\mathrm{DVS}_{\mathrm{avg}@32}$: the mean Data Verifier Score across all 32 recipes. This metric quantifies the expected quality and stability of the policy, where recipes failing to yield valid training data are assigned a score of $0$.
(2) $\mathrm{DBS}$: the Downstream Benchmark Score of a model trained on a single recipe, which is randomly sampled from the subset of candidates with valid execution (i.e., $\mathrm{DVS}>0$). This metric reflects the actual performance on the downstream benchmark of a successfully executed recipe.
Additionally, to approximate the oracle upper bound for DataChef-32B, we select the most promising recipe from the candidate set and report its downstream score.
For all downstream evaluation, we fine-tune Qwen3-1.7B-Base for 3 epochs with a learning rate of 2e-5 and a batch size of 64.

\noindent
\textbf{Baselines.}
We compare DataChef-32B against leading LLMs, including Qwen3-32B, Kimi-K2-Instruct~\cite{kimik2}, Qwen3-Next-80B-A3B-Thinking, and Gemini-3-Pro~\cite{gemini-3}. Additionally, we incorporate the following results as reference:
(1) To benchmark raw data quality and hand-designed strategies, we manually filter and format each available source, reporting the average and best downstream performance trained on single-source datasets. Furthermore, we apply SOTA data selection algorithms, IFD~\cite{IFD} and DEITA~\cite{Deita}, to this manually pre-processed pool.
(2) We report the performance of the official Qwen3-1.7B checkpoint, representing an industry-standard topline achieved via expert-curated data recipes.
Detailed experiment setups are provided in Appx.~\ref{appx:exp-setup}.

\subsection{Main Results}
\label{sec:main-results}
\noindent
\textbf{Main Comparison.} 
Table~\ref{tab:main-results} presents the performance of DataChef-32B against baselines across in-domain and out-of-domain tasks.
DataChef-32B achieves superior performance compared to a strong practical baseline, Qwen3-Next-80B $\oplus$ Kimi-K2, which leverages open-source state-of-the-art specialized models (Qwen3-Next-80B-A3B-Thinking for reasoning and Kimi-K2-Instruct for coding).
Specifically, our end-to-end model surpasses this composite system with average improvements of +8.6\% and +9.2\% in $\mathrm{DVS}_{\mathrm{avg}@32}$, and +10.7\% and +7.4\% in $\mathrm{DBS}$ on in-domain and out-of-domain tasks, respectively.
Notably, DataChef-32B achieves performance comparable to the closed-source top-tier Gemini-3-Pro, demonstrating exceptional robustness and effectiveness in automated data recipe generation.

\begin{figure}[t]
    \centering
    \includegraphics[width=0.98\linewidth]{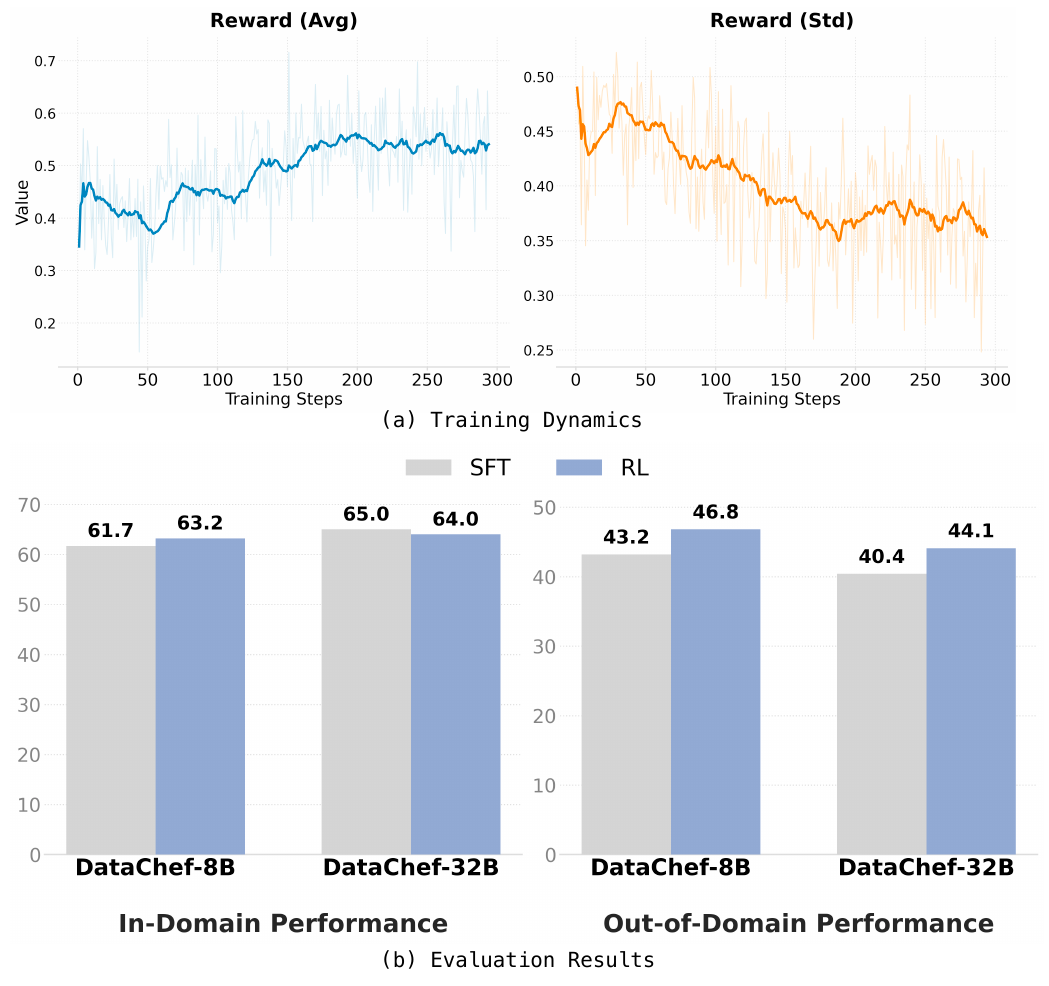}
    \vspace{-4pt}
    \caption{\textbf{Analysis of RL Effectiveness}. (a) RL training dynamics indicate that the policy consistently converges toward high-quality data recipe generation. (b) Evaluation results show that RL yields substantial improvements on out-of-domain tasks.}
    \label{fig:rl-eff}
    \vspace{-16pt}
\end{figure}

\begin{table}[t]
\centering
\footnotesize
\caption{\textbf{Ablation study on training stages and reward design}. We investigate the impact of the cold-start phase and the granularity of the reward signal. $\mathcal{R}_{\text{dense}}$ denotes our proposed fine-grained Data Verifier score, while $\mathcal{R}_{\text{sparse}}$ represents a constant success reward for valid execution.}
\label{tab:abla}
 \scalebox{0.68}{
\begin{tabular}{l ccc cc}
\toprule
\multirow{2}{*}{\textbf{Model}} & \multirow{2}{*}{\textbf{Cold Start}} & \multirow{2}{*}{\textbf{RL}} & \multirow{2}{*}{\textbf{Reward}} & \multicolumn{2}{c}{\textbf{Performance} ($\mathrm{DVS}_{\mathrm{avg}@32}$)} \\
\cmidrule(lr){5-6}
& & &  & \textbf{In-Domain} & \textbf{Out-of-Domain} \\
\midrule
$\mathcal{M}_{Baseline}$ & $\times$ & $\times$ & - & 4.1 & 5.5 \\
$\mathcal{M}_{RL}$       & $\times$ & $\checkmark$ & $\mathcal{R}_{\text{dense}}$ & 32.9 & 23.9 \\
$\mathcal{M}_{Sparse}$   & $\checkmark$ & $\checkmark$ & $\mathcal{R}_{\text{sparse}}$ & 62.7 & 44.1 \\
\rowcolor{MyGray} DataChef-8B & $\checkmark$ & $\checkmark$ & $\mathcal{R}_{\text{dense}}$ & 63.2 & 46.8 \\
\bottomrule
\end{tabular}
}
\vspace{-4pt}
\end{table}

\begin{table}[t]
\centering
\small
\caption{\textbf{Analysis of collaborating with strong coding models}. We compare the end-to-end paradigm against decoupled approaches where the model acts solely as a planner, relying on an external coder (Kimi-K2-Instruct) for implementation.}
\scalebox{0.74}{
\label{tab:coder}

\begin{tabular}{l c cc}
\toprule
\multirow{2}{*}{\textbf{Model}} & \multirow{2}{*}{\textbf{External Coder}} & \multicolumn{2}{c}{\textbf{Performance} ($\mathrm{DVS}_{\mathrm{avg}@32}$)} \\
\cmidrule(lr){3-4}
&  & \textbf{In-Domain} & \textbf{Out-of-Domain} \\
\midrule
\rowcolor{gray!10} \multicolumn{4}{l}{\textit{\textbf{Inference-Time}}} \\
Qwen3-32B & $\times$ & 22.3 & 18.9 \\
Qwen3-32B & \checkmark & 40.3 & 33.1 \\
\midrule
\rowcolor{gray!10} \multicolumn{4}{l}{\textit{\textbf{Training Paradigm}}} \\
Planner-32B  & \checkmark & 56.7 & 37.3 \\
\textbf{DataChef-32B} & $\times$ & \textbf{64.0} & \textbf{44.1} \\
\bottomrule
\end{tabular}
}
\vspace{-8pt}
\end{table}

\noindent
\textbf{Surpassing Human Baselines.}
By selecting the most promising recipe from 32 samples (Oracle Upper Bound), DataChef-32B outperforms $\textsc{Single-Source}_{\mathrm{best}}$, IFD, and DEITA across most tasks, achieving an average in-domain score of 130.3.
This indicates that DataChef goes beyond simple dataset selection and synthesizes novel data processing pipelines, including effective selection, mixing, synthesis, and filtering, thereby demonstrating the advantage of automatically exploring a vast code space over human-designed heuristics.
Remarkably, it achieves 66.7 on AIME'25 and 46.3 on ClimaQA, surpassing the official checkpoint trained with industry-level, expert-curated data recipes.
These results underscore the potential of fully automating data recipe generation for LLM training.

\subsection{Data Verifier}
\label{sec:data-verifier}
\noindent
\textbf{Setup.}
To validate the proposed Data Verifier, we analyze the Pearson correlation between the verifier scores and downstream benchmark performance.
We benchmark against several widely used data evaluation metrics, including IFD~\cite{IFD}, RewardModelScore~\cite{skywork}, DEITA~\cite{Deita}, and VendiScore~\cite{VendiScore}.
To ensure diversity in data quality and model performance, we construct 8--12 datasets per task under a fixed data budget using two strategies:
(1) Direct sampling from available task-specific data sources.
(2) Subset selection from the pool formed in (1) based on response length.

\noindent
\textbf{Correlation Analysis.}
As shown in Fig.~\ref{fig:cor-analysis} and detailed in Appx.~\ref{appx:additional-results-on-correlation-analysis}, the Data Verifier exhibits superior capability compared to existing metrics, achieving the highest average Pearson correlation of $0.59$ across all six domains.
Crucially, while baseline metrics suffer from severe cross-domain variance, frequently yielding negative correlations that can provide misleading optimization signals in domains such as Math ($-0.48$ for IFD) and Code ($-0.15$ for DEITA), our Data Verifier maintains a strictly positive correlation across all evaluated tasks.
Furthermore, it demonstrates the highest statistical significance, with 4 out of 6 tasks satisfying $p < 0.1$, confirming that the Data Verifier provides a statistically robust and globally consistent signal for automated data recipe generation.


\begin{figure}[t]
    \centering
    \includegraphics[width=0.98\linewidth]{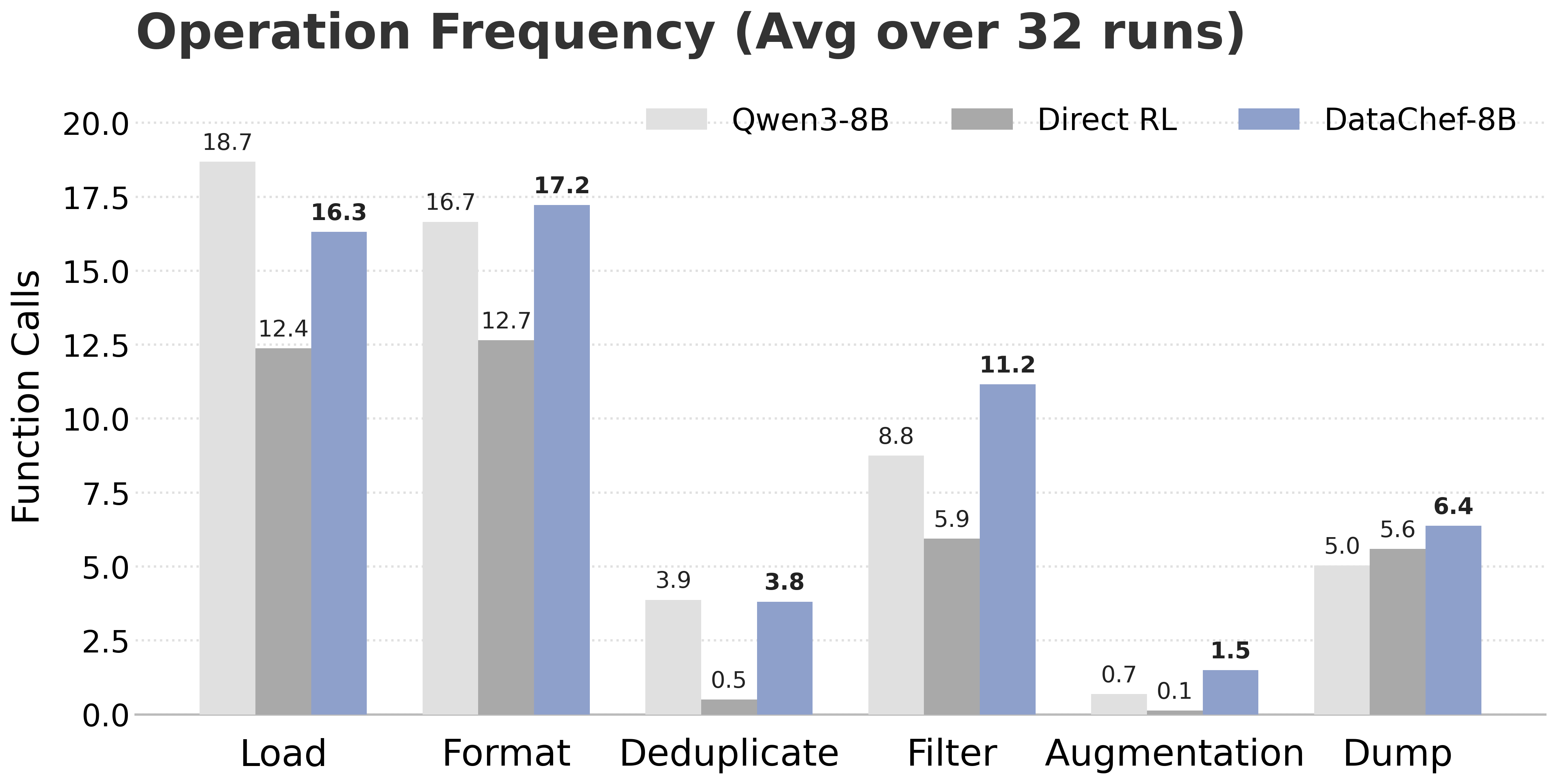}
    \caption{\textbf{Analysis of operation frequency in generated recipes}. We compare the average number of function calls per recipe across different models.}
    \label{fig:op-fre}
    \vspace{-12pt}
\end{figure}

\subsection{Ablation and Analysis}

\noindent
\textbf{Effectiveness of RL.}
Fig.~\ref{fig:rl-eff} illustrates that reward values consistently trend upward while the standard deviation decreases during training, confirming the convergence and effectiveness of RL process.
Held-out evaluation reveal that RL primarily enhances generalization, yielding significant gains on out-of-domain tasks while preserving in-domain performance.
Quantitatively, RL delivers an average $\mathrm{DVS}_{\mathrm{avg}@32}$ improvement of 3.6\% for the 8B model and 3.7\% for the 32B model.

\noindent
\textbf{Effectiveness of Cold Start.}
Table.~\ref{tab:abla} shows that omitting the cold start leads to significant performance degradation across all domains.
To understand this behavior, we analyze the distribution of function calls in Fig.~\ref{fig:op-fre}.
The results demonstrate that the direct RL model tends to generate simplistic data pipelines, reducing the usage of complex data processing operations.
We hypothesize that without the SFT warm-up, the model succumbs to \textit{reward hacking}. It avoids execution penalties by generating safe, trivial scripts rather than optimizing for data quality.
In contrast, DataChef-8B leverages the SFT foundation to explore and deploy sophisticated operations, such as filtering and data augmentation.

\noindent
\textbf{Ablation on Reward Signal.}
To assess the effectiveness of the fine-grained data quality feedback, we conduct an ablation where the continuous verifier score $s(\hat{d})$ in Eq.~\ref{eq:reward} is replaced by a constant success reward (i.e., assigning a fixed value of $1.0$ to any valid data recipe).
Table~\ref{tab:abla} demonstrates that this quality-agnostic signal leads to noticeable performance drops.
This result confirms that the model relies on the guidance from the Data Verifier to distinguish high-utility recipes from merely executable ones.

\noindent
\textbf{Collaborating with Strong Coder.}
\label{sec:coll}
Given the proliferation of specialized coding models, a natural idea is to decouple this task: use the primary model as a planner (natural language orchestration) and an external coder for implementation.
Table~\ref{tab:coder} shows that this paradigm enhances inference-time performance, with Qwen3-32B paired with Kimi-K2-Instruct yielding 18.0\% and 14.5\% $\mathrm{DVS}_{\mathrm{avg}@32}$ gains on in-domain and out-of-domain tasks, respectively.
However, training the model solely as a planner leads to suboptimal results compared to the end-to-end approach.
This suggests that integrated training of planning and coding capabilities is essential for optimal data recipe generation.

\begin{figure}[t]
    \centering
    \includegraphics[width=0.9\linewidth]{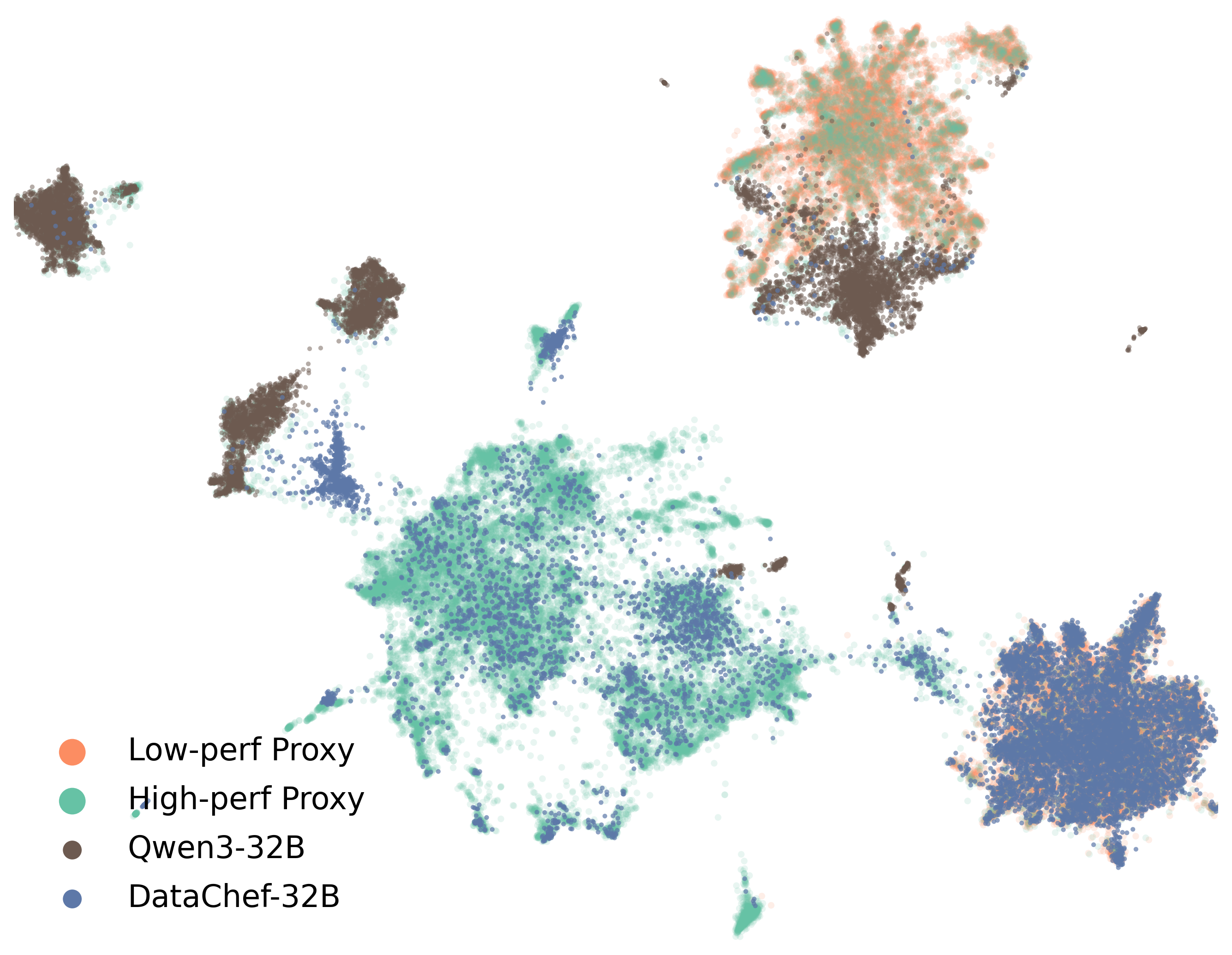}
    \caption{\textbf{Visualization of data distribution in generated recipes}. We project the source datasets and the data recipes generated by different models into a 2D embedding space.}
    \label{fig:fin-case}
    \vspace{-12pt}
\end{figure}

\noindent
\textbf{Case Study.}
We quantitatively analyze the data recipes generated by Qwen-32B and DataChef-32B for the out-of-domain financial task in Fig~\ref{fig:fin-case}.
We categorize source datasets that yield high downstream performance as \textit{High-perf proxy}, and those performing poorly as \textit{Low-Perf Sources}.
DataChef-32B demonstrates an emergent ability to identify and prioritize high-utility datasets.
Additionally, we provide detailed data processing pipelines with code examples in Appx.~\ref{appx:case-study}, which reveals that DataChef-32B can: (1) automatically leverage LLMs to augment data into task-specific formats or to synthesize data to enhance target ability; and (2) extract the most relevant data subsets using self-generated keywords.
\section{Conclusion}
In this paper, we propose a novel paradigm for automated data recipe generation to streamline LLM adaptation. To facilitate this, we establish a holistic dataset for both training and evaluation. Building on this foundation, we present DataChef-32B, incorporating a data verifier that serves as a cost-effective reward function for online RL. 
DataChef-32B demonstrates strong generalization capabilities, matching human-level expertise on specific benchmarks. Our work bridges the gap between data curation and model evolution, fostering the development of self-evolving AI.

\noindent
\textbf{Limitations.}
Our reliance on an LLM-as-a-Judge for proxy rewards prioritizes generalizability but may sacrifice precision in niche tasks. Developing specialized evaluators to offer higher-resolution reward signals remains a valuable direction for future research.

\bibliography{main}

\clearpage
\appendix

\section{Implementation Details of DataChef}

\subsection{Details of Task Pool}
\label{appx:task-pool}

\noindent
\textbf{Full Benchmark List.}
Table~\ref{tab:benchmark_details} presents the full list of benchmarks used in our task pool, along with their domains and usage.

\begin{table}[h] 
    \centering 
    \footnotesize 
    \caption{List of benchmarks used in the task pool.} 
    \label{tab:benchmark_details}  
    \scalebox{0.72}{ 
    \begin{tabular}{l l c} 
        \toprule 
        \textbf{Domain} & \textbf{Benchmark} & \textbf{Usage} \\ 
        \midrule

        \multirow{2}{*}{Code} & HumanEval~\cite{humaneval} & Train \\
                              & LiveCodeBench v6~\cite{livecodebench} & Test \\
        
        \midrule

        \multirow{2}{*}{Comprehension} & OpenbookQA~\cite{openbookqa} & Train \\
                                       & RACE~\cite{race} & Train \\
        \midrule

        Instruction Following & IFEval~\cite{ifeval} & Train \\
        \midrule

        \multirow{2}{*}{Agriculture} & AgXQA~\cite{agxqa} & Train \\
                                     & SeedBench~\cite{seedbench} & Train \\
        \midrule 

        Astronomy & Astrobench~\cite{astrobench} & Train \\
        \midrule

        \multirow{3}{*}{Biology} & Genome-Bench~\cite{genomebench} & Train \\
                                 & MoleculeQA~\cite{moleculeqa} & Train \\
                                 & ProteinLMBench~\cite{proteinlmbench} & Train \\
        \midrule

        Chemistry & ChemBench~\cite{chembench} & Train \\
                                   
        \midrule

        Earth Science & EarthSE~\cite{earthse} & Train \\
        \midrule

        \multirow{2}{*}{General Knowledge} & C-Eval~\cite{ceval} & Train \\
                                           & MMLU~\cite{mmlu} & Train \\
        \midrule

        \multirow{2}{*}{Medical} & MedXpertQA~\cite{medxpertqa} & Train \\
        & MedQA~\cite{medqa} & Train \\
        \midrule

        Finance & OpenFinData~\cite{openfindata} & Test \\
        \midrule

        Atmosphere & ClimaQA~\cite{climaqa} & Test \\
        \midrule

        \multirow{2}{*}{Physics} & PHYBench~\cite{phybench} & Train \\
                                 & PHYSICS~\cite{physics} & Test \\
        \midrule

        \multirow{2}{*}{Long Context} & L-Eval~\cite{leval} & Train \\
                                      & LongBench~\cite{longbench} & Train \\
        \midrule

        \multirow{2}{*}{Math} & GSM8K~\cite{gsm8k} & Train \\
        & AIME'25~\cite{aime} & Test \\
        
        \midrule

        \multirow{2}{*}{Reasoning} & BBH~\cite{bbh} & Train \\
                                   & HotpotQA~\cite{hotpotqa} & Train \\
        \midrule

        \multirow{2}{*}{Safety} & HaluEval~\cite{HaluEval} & Train \\
                                & SafetyBench~\cite{safetybench} & Train \\
        \midrule

        Writing & WritingBench~\cite{writingbench} & Train \\

        \midrule

        Language & CHID~\cite{chid} & Test \\

        \bottomrule 
    \end{tabular} 
    } 
\end{table}

\noindent
\textbf{Datasets Retrieval Procedure.}
We automate the retrieval of candidate datasets relevant to the benchmark and task through the following procedure:
\begin{enumerate}[label={\bf {{$\bullet$}}}, leftmargin=*, topsep=0.5ex, itemsep=0ex, wide, labelindent=0pt]
    \item \textbf{Keyword Synthesis.}  We use an LLM to generate 3--5 high-relevance search keywords tailored to the downstream task.
    \item \textbf{Search.} Leveraging the generated keywords, we query both the Hugging Face and Google Search APIs to harvest a broad spectrum of potential datasets.
    \item \textbf{Rank.} To ensure dataset utility, we rank retrieved datasets by community popularity (e.g., likes) and retain the top-4 candidates for each keyword.
    \item \textbf{Verify.} We implement a strict verification protocol to preclude data leakage, explicitly ensuring no overlap exists between the retrieved candidates and the benchmark.
\end{enumerate}

\subsection{Prompt Templates and Model Selection}
\label{appx:prompt}

\begin{figure*}[t]
\begin{tcolorbox}[colback=gray!5!white,colframe=gray!75!black,title=\textbf{Prompt for Data Verifier}]
\small
As a grading expert, your task is to determine whether the candidate's response matches the question and to assess the sample's usefulness for the specified task. Follow these evaluation guidelines precisely:

\textbf{Evaluation Protocol:}

\textbf{1. Validity Check:}
\begin{itemize}[leftmargin=*, nosep]
    \item \textbf{Reject questions} that are: INCOMPLETE (cut off), REPETITIVE (loops), or NOT\_ENOUGH\_INFO.
    \item \textbf{Reject answers} that are: INCOMPLETE, REPETITIVE, REFUSAL (e.g., "I cannot answer..."), or IRRELEVANT.
    \item \textbf{Action}: Classify as \texttt{\textbackslash boxed\{A\}} and specify the reason (e.g., \texttt{\textbackslash boxed\{A\} - INCOMPLETE}).
\end{itemize}

\textbf{2. Format Check:}
\begin{itemize}[leftmargin=*, nosep]
    \item Ensure the answer follows any explicit output-format requirements (e.g., single choice, JSON schema).
    \item If no explicit format is required, pass.
    \item \textbf{Action}: If violated, classify as \texttt{\textbackslash boxed\{B\} - FORMAT\_ERROR}.
\end{itemize}

\textbf{3. Correctness Check:}
\begin{itemize}[leftmargin=*, nosep]
    \item Re-generate a concise reference answer and compare it with the candidate's final answer.
    \item For multi-part questions, require all parts to be correct; partial correctness $\rightarrow$ Fail.
    \item \textbf{Action}: If mismatched, classify as \texttt{\textbackslash boxed\{C\} - INCORRECT}.
\end{itemize}

\textbf{4. Task-Alignment Check (Training-Suitability):}
\begin{itemize}[leftmargin=*, nosep]
    \item Evaluate if the sample (Q+A) is useful for training the specified task.
    \item \textbf{Scope Fit}: Targets the capabilities the task trains.
    \item \textbf{I/O Contract Impact}:
    \begin{itemize}[leftmargin=10pt, nosep]
        \item \textit{Beneficial/Benign (ALIGN)}: Mild deviations that still teach the target mapping (e.g., adding a brief rationale where not strictly forbidden).
        \item \textit{Fatal Mismatch (MISMATCH)}: Conflicts likely to cause inference failure (e.g., task requires JSON but sample invents schema; task requires single letter but sample gives long essay without choice).
    \end{itemize}
    \item \textbf{Action}: If harmful noise or fatal I/O mismatch $\rightarrow$ \texttt{\textbackslash boxed\{D\} - TASK\_MISMATCH}. Else $\rightarrow$ \texttt{\textbackslash boxed\{E\} - PASS}.
\end{itemize}

\textbf{Grading Scale:}
\begin{itemize}[leftmargin=*, nosep]
    \item \texttt{\textbackslash boxed\{A\} - INVALID}: Fails validity criteria.
    \item \texttt{\textbackslash boxed\{B\} - FORMAT\_ERROR}: Fails format check.
    \item \texttt{\textbackslash boxed\{C\} - INCORRECT}: Deviates from reference answer.
    \item \texttt{\textbackslash boxed\{D\} - TASK\_MISMATCH}: Fails task-alignment check.
    \item \texttt{\textbackslash boxed\{E\} - PASS}: Passes all checks.
\end{itemize}

\textbf{Execution Steps:}

1. Thoroughly evaluate validity, format, correctness, and task-alignment step-by-step.

2. If any check fails, immediately assign the corresponding grade.

\textbf{Input Data:}

<Task Description Begin> 

\{\{ task\_description \}\} 

<Task Description End>

<Original Question Begin> 

\{\{ question \}\} 

<Original Question End>\\
<Original Answer Begin> 

\{\{ llm\_response \}\} 

<Original Answer End>

\textbf{Output Format:}
\texttt{Analysis step by step: [...]}\\
\texttt{Final Judgment: \textbackslash boxed\{GRADE\} - REASON}
\end{tcolorbox}
\caption{Prompt used for Data Verifier.}
\label{fig:verifier-prompt}
\end{figure*}

\noindent
\textbf{Data Verifier.}
We utilize gpt-oss-120b~\cite{gptoss} as the Data Verifier. This is a Mixture-of-Experts (MoE) model with only about 5B active parameters per token, ensuring high inference speed.
The detailed rubric-based prompt used for evaluation is presented in Fig.~\ref{fig:verifier-prompt}.

\begin{figure*}[t]
\begin{tcolorbox}[colback=gray!5!white,colframe=gray!75!black,title=\textbf{Prompt for Data Recipe Generation (Natural Language)}]
\small
\texttt{\# Task Description}\\
\texttt{\{\{ task\_description \}\}}

\texttt{\# Benchmark}\\
\texttt{\#\# \{\{ benchmark.name \}\}}\\
\texttt{\{\{ benchmark.description \}\}}

\texttt{\# Available Hugging Face Training Datasets}\\
\texttt{\{\% for item in datasets -\%\}}\\
\texttt{\#\# \{\{ item.dataset\_id \}\}}\\
\texttt{\{\{ item.examples \}\}}\\
\texttt{\{\% endfor -\%\}}

---

Based on the Task Description, the target Benchmark, and the Available Hugging Face Training Datasets, design a feasible \textbf{Data Processing Plan}. This plan must include:
(1) **Data Selection**: Identify the most suitable raw datasets from the available list.
(2) **Data Processing Workflow**: Define the pipeline to transform selected raw data into high-quality SFT training data.

The plan will serve as a blueprint for code generation and data production. Ensure it is comprehensive, actionable, and free from ambiguous or vague statements.

A high-quality SFT dataset must exhibit the following attributes:
\begin{itemize}[leftmargin=*]
    \item \textbf{High Quality}: Accurate samples free from noise.
    \item \textbf{Diversity}: Coverage of varied instructions and objectives.
    \item \textbf{Relevance}: Strong alignment with the target vertical domain.
\end{itemize}

\textbf{Important Constraints \& Guidelines:}
\begin{itemize}[leftmargin=*]
    \item \textbf{Source Selection}: strictly select datasets provided in the context. Prioritize those with complete metadata and clear field definitions. \textbf{Strictly prohibit data contamination: DO NOT use benchmark data for training.} Do not hallucinate datasets, splits, or configurations not provided in the context.
    \item \textbf{Grounding}: Base the processing workflow solely on the actual fields and content of the selected datasets. Do not make assumptions about data fields that do not exist.
    \item \textbf{Context}: Ensure inputs and outputs form coherent Q\&A turns. If the output relies on specific context (e.g., a document or snippet) within the data, design the workflow to explicitly embed this context into the input.
    \item \textbf{Format Alignment}: If the benchmark requires specific formats (e.g., multiple-choice, JSON output), ensure the constructed training data aligns with these requirements.
    \item \textbf{Final Output Format}: The pipeline must produce data in the following standard dialogue format:\\
    \texttt{\{"dialogs": [\{"role": "user", "content": "..."\}, \{"role": "assistant", "content": "..."\}]\}}
    \item \textbf{LLM Utilization}: Flexibly leverage LLM inference capabilities for tasks such as: extracting instructions from heterogeneous documents, data augmentation/synthesis, and quality verification. Clearly specify the prompt design strategies for these steps.
\end{itemize}

\textbf{Output Format:}
Generate the plan strictly in the following format (do not include preamble or additional text):

\texttt{\#\# Training Data}\\
\texttt{[}\\
\texttt{    \{"dataset\_id": "target\_id", "split": "target\_split", "name": "config\_name", "sample\_num": "int", "reason": "justification"\},}\\
\texttt{    ...}\\
\texttt{]}

\texttt{\#\# Data Processing Workflow}\\
\texttt{[Detailed Data Processing Workflow description]}
\end{tcolorbox}

\caption{Prompt used for data recipe generation (Natural Language).}
\label{fig:planner-prompt}
\end{figure*}

\begin{figure*}[t]
\begin{tcolorbox}[colback=gray!5!white,colframe=gray!75!black,title=\textbf{Prompt for Data Recipe Generation (Executable Code)}]
\small
\texttt{\# Available Hugging Face Training Datasets}\\
\texttt{\{\% for item in datasets -\%\}}\\
\texttt{\#\# \{\{ item.dataset\_id \}\}}\\
\texttt{\{\{ item.examples \}\}}\\
\texttt{\{\% endfor -\%\}}

\texttt{\# Data Processing Plan}\\
\texttt{\{\{ plan \}\}}

\texttt{\# Tool Information}\\
\texttt{\{\{ tool\_info \}\}}

---

Based on the Available Hugging Face Training Datasets, the Data Processing Plan, and the Tool Information, generate the executable **Data Processing Script**.

To validate the correctness of the processing, also generate a corresponding **Verification Script**.

\textbf{Important Notes \& Constraints:}
\begin{itemize}[leftmargin=*]
    \item \textbf{Implementation Logic}: The script must clearly implement the processing rationale and utilize tools correctly.
    \item \textbf{Verification Scope}: The verification script should validate that the generated data matches expectations and confirm the effectiveness of core processing steps.
    \item \textbf{Target Format}: The final data must follow the ShareGPT format:\\ \texttt{\{'dialogs': [{'role': 'user', 'content': '...'}, {'role': 'assistant', 'content': '...'}]\}}. Use the \texttt{format\_to\_sharegpt} utility (import from \texttt{aidp}).
    \item \textbf{Output Directory}: Save the processed data to the \texttt{data/processed/} directory.
\end{itemize}

\textbf{Output Format:}
Generate exactly two Python code blocks in the following format, without any additional text:

\texttt{```python}\\
\texttt{\# data-processing code block}\\
\texttt{```}

\texttt{```python}\\
\texttt{\# test code block}\\
\texttt{```}
\end{tcolorbox}
\caption{Prompt used for data recipe generation (Executable Code).}
\label{fig:coder-prompt}
\end{figure*}

\noindent
\textbf{Cold-start Models.}
To construct high-quality cold-start supervision, we employ two specialized models: Qwen3-Next-80B-A3B-Thinking~\cite{qwen3} for planning and reasoning, and Kimi-K2-Instruct~\cite{kimik2} for code implementation. The corresponding prompts are detailed in Fig.~\ref{fig:planner-prompt} and Fig.~\ref{fig:coder-prompt}.

\section{Details of Experiments Setup}
\label{appx:exp-setup}

\subsection{Data Evaluation Metrics Settings}
We use the OpenDataArena-Tool\footnote{\url{https://github.com/OpenDataArena/OpenDataArena-Tool}} for data assessment, adhering to its default configurations.
The specific settings for the data evaluation metrics used in our experiments are as follows:
\begin{enumerate}[label={\bf {{$\bullet$}}}, leftmargin=*, topsep=0.5ex, itemsep=0ex, wide, labelindent=0pt]
    \item \textbf{IFD.} We employ Qwen2.5-3B-Instruct as the backend model to calculate the Instruction-Following Difficulty (IFD) score. Following~\cite{IFD}, instances with an IFD score $> 1$ are treated as outliers. To ensure robust correlation analysis, we assign a score of $0$ to these anomalies.
    \item \textbf{DEITA.} Following~\cite{Deita}, we define the final data score as the product of the \textit{Complexity Score} and the \textit{Quality Score}. These scores are computed using the checkpoints provided in the official DEITA repository.
    \item \textbf{RewardModelScore.} We utilize Skywork-Reward-V2-Llama-3.1-8B-40M~\cite{skywork} to compute the reward score, serving as a proxy for response quality.
    \item \textbf{VendiScore.} We employ Qwen3-Embedding-0.6B to compute sample embeddings and utilize Euclidean distance as the similarity metric to calculate VendiScore, measuring the diversity of the dataset.
\end{enumerate}

\begin{table*}[t]
\centering
\caption{\textbf{Detailed correlation results across different domains}. For each domain, we report the Pearson correlation coefficient ($r$) and the corresponding $p$-value.}
\label{tab:correlation_results}
\scalebox{0.72}{
\begin{tabular}{lccccccccccccccccc}
\toprule
\multirow{2}{*}{\textbf{Metric}} & \multicolumn{2}{c}{\textbf{PHYSICS}} & \multicolumn{2}{c}{\textbf{AIME}} & \multicolumn{2}{c}{\textbf{LiveCodeBench}} & \multicolumn{2}{c}{\textbf{ClimaQA}} & \multicolumn{2}{c}{\textbf{OpenFinData}} & \multicolumn{2}{c}{\textbf{CHID}} & \multicolumn{3}{c}{\textbf{Summary}} \\
\cmidrule(lr){2-3} \cmidrule(lr){4-5} \cmidrule(lr){6-7} \cmidrule(lr){8-9} \cmidrule(lr){10-11} \cmidrule(lr){12-13} \cmidrule(lr){14-16}
& $r$ & $p$ & $r$ & $p$ & $r$ & $p$ & $r$ & $p$ & $r$ & $p$ & $r$ & $p$ & Positive & $p<0.05$ & $p<0.1$ \\
\midrule
IFD          & 0.48 & 0.19 & -0.48 & 0.12 & 0.43 & 0.25 & 0.90 & 0.00 & 0.59 & 0.10 & 0.51 & 0.16 & 5/6 & 1/6 & 2/6 \\
Deita        & 0.20 & 0.61 & -0.04 & 0.91 & -0.15 & 0.70 & 0.91 & 0.00 & 0.10 & 0.80 & 0.61 & 0.08 & 4/6 & 1/6 & 2/6 \\
RewardModel  & 0.34 & 0.37 & 0.18 & 0.58 & 0.30 & 0.43 & 0.90 & 0.00 & 0.36 & 0.34 & 0.44 & 0.23 & 6/6 & 1/6 & 1/6 \\
VendiScore   & 0.36 & 0.34 & 0.33 & 0.29 & -0.17 & 0.67 & 0.66 & 0.08 & 0.60 & 0.09 & 0.29 & 0.44 & 5/6 & 0/6 & 2/6 \\
DataVerifier & 0.63 & 0.07 & 0.59 & 0.04 & 0.63 & 0.07 & 0.52 & 0.19 & 0.34 & 0.37 & 0.83 & 0.01 & 6/6 & 2/6 & 4/6 \\
\bottomrule
\end{tabular}
}
\end{table*}

\subsection{Evaluation Setup}
\label{appx:evaluation-setup}
All downstream task evaluations are conducted using the OpenCompass framework\footnote{\url{https://github.com/open-compass/opencompass}}. The detailed settings for each benchmark are as follows:
\begin{enumerate}[label={\bf {{$\bullet$}}}, leftmargin=*, topsep=0.5ex, itemsep=0ex, wide, labelindent=0pt]
    \item \textbf{PHYSICS.} We employ xVerify-9B-C~\cite{xverify} as the evaluator and report the average accuracy across all sub-tasks.
    \item \textbf{AIME'25.} We evaluate on the 2025 subset (covering both Part I and Part II). For each question, we generate 8 responses and report the average accuracy. xVerify-9B-C is used as the evaluator.
    \item \textbf{LiveCodeBench v6.} We utilize the official prompt guidelines and report the pass@1 metric. The LCBCGenerationEvaluator is used for assessment.
    \item \textbf{ClimaQA.} We employ xVerify-9B-C as the evaluator and report the average accuracy across all sub-tasks.
    \item \textbf{OpenFinData.} We use the OpenFinDataKWEvaluator and report the average accuracy across all sub-tasks.
    \item \textbf{CHID.} We report the average accuracy on both the development and test sets.
\end{enumerate}

\subsection{Oracle Selection}
\label{appx:oracle-selection}
The specific procedure for Oracle Selection is:
\begin{enumerate}[label={}, leftmargin=*, topsep=0.5ex, itemsep=0ex, wide, labelindent=0pt]
    \item \textbf{1. Filter:} From the 32 generated recipes, we select the top-8 candidates based on the Data Verifier Score.
    \item \textbf{2. Verify:} A human expert reviews these top-8 candidates to ensure quality, verifying key aspects including: (a) format alignment with the target benchmark (e.g., the processed dataset should include multiple-choice structures for MCQ tasks); (b) context integrity (e.g., verifying that inputs include necessary reference texts for reading comprehension, rather than just questions); and (c) logical comprehensiveness of the pipeline operations. Based on this comprehensive assessment, the expert selects the optimal candidate for final training.
\end{enumerate}

\subsection{Budget}
\label{appx:budget}
For DBS evaluation, the data budget is set to 10K. If a recipe generates more, we randomly downsample to 10K; if fewer, we use all. During training and DVS evaluation, we use a budget of 100 samples for rapid proxy reward calculation.

\section{Case Study}
\label{appx:case-study}
To demonstrate the capability of our model, we present a complete data processing pipeline generated by DataChef-32B for the ClimaQA task. As shown in Fig.~\ref{fig:code-example}, the generated code successfully produces valid training data by: (1) automatically leveraging LLMs to augment data into task-specific formats and synthesize samples to enhance target capabilities; and (2) extracting the most relevant data subsets using self-generated keywords.

\begin{figure*}[h]
\centering
\begin{lstlisting}[language=Python, title={Generated Data Recipe Code (ClimaQA Case Study Part I)}]

# data-processing code block
...

# 1. Load stackexchange_earthscience
stack_ds = load_remote_dataset(
    "mlfoundations-dev/stackexchange_earthscience",
    name="default",
    split="train",
    shuffle=False
)

# 1.1 Use LLM to convert Open QA into 4-option Multiple Choice Questions
SYSTEM_PROMPT_STACK = """
You are a science tutor.  
Convert the given open QA pair into a 4-option multiple-choice question.  
Return JSON only:
{"question": "<original question>", "A": "<option A>", "B": "<option B>", "C": "<option C>", "D": "<option D>", "answer": "<correct letter>"}
The correct answer must be one of the four options. Distractors should be scientifically plausible but wrong.
"""

def stack_parser(response: str, raw: dict) -> list:
    parsed = extract_json(response)
    if not parsed or "answer" not in parsed:
        return []
    opts = f"A) {parsed['A']}\nB) {parsed['B']}\nC) {parsed['C']}\nD) {parsed['D']}"
    user = f"Question: {parsed['question']}\nOptions:\n{opts}"
    assistant = parsed["answer"].lower()
    return [{"user": user, "assistant": assistant}]

stack_mc = generate_dataset_with_llm(
    stack_ds,
    system_prompt=SYSTEM_PROMPT_STACK,
    response_parser=stack_parser
)
\end{lstlisting}
\end{figure*}

\begin{figure*}[h]
\centering
\begin{lstlisting}[language=Python, title={Generated Data Recipe Code (ClimaQA Case Study Part II)}]

# 2. Load ScienceQA_text_only 
sciqa_ds = load_remote_dataset(
    "tasksource/ScienceQA_text_only",
    name="default",
    split="train",
    shuffle=False
)

# 2.1 Filter for climate-related topics
def climate_filter(sample):
    txt = (sample.get("category", "") + " " + sample.get("topic", "")).lower()
    return any(k in txt for k in ("climate", "weather", "atmosphere", "ocean", "environment"))

sciqa_climate = select_by_filter(sciqa_ds, climate_filter)

# 2.2 Standardize format
def sciqa_map(sample):
    choices = sample["choices"]
    opts = "\n".join([f"{chr(65+i)}) {c}" for i, c in enumerate(choices)])
    user = f"Question: {sample['question']}\nOptions:\n{opts}"
    assistant = chr(65 + sample["answer"])
    return {"user": user, "assistant": assistant}

sciqa_mc = sciqa_climate.map(sciqa_map, remove_columns=sciqa_climate.column_names)

# 3. Merge & Deduplicate
merged = concatenate_datasets([stack_mc, sciqa_mc])
deduped = deduplicate_by_text_hash(
    merged,
    text_map=lambda x: x["user"],
    lowercase=True,
    ignore_non_character=True
)

# 4. Convert to ShareGPT format
sharegpt_ds = format_to_sharegpt(
    deduped,
    user_map=lambda x: x["user"],
    assistant_map=lambda x: x["assistant"]
)

# 5. Save output
dump_dataset(sharegpt_ds, "data/processed/train_climaqa_style.jsonl")
\end{lstlisting}
\caption{Case study of data recipe generation.}
\label{fig:code-example}
\end{figure*}

\section{Additional Results on Correlation Analysis}
\label{appx:additional-results-on-correlation-analysis}
\begin{figure*}[t]
    \centering
    \includegraphics[width=0.95\linewidth]{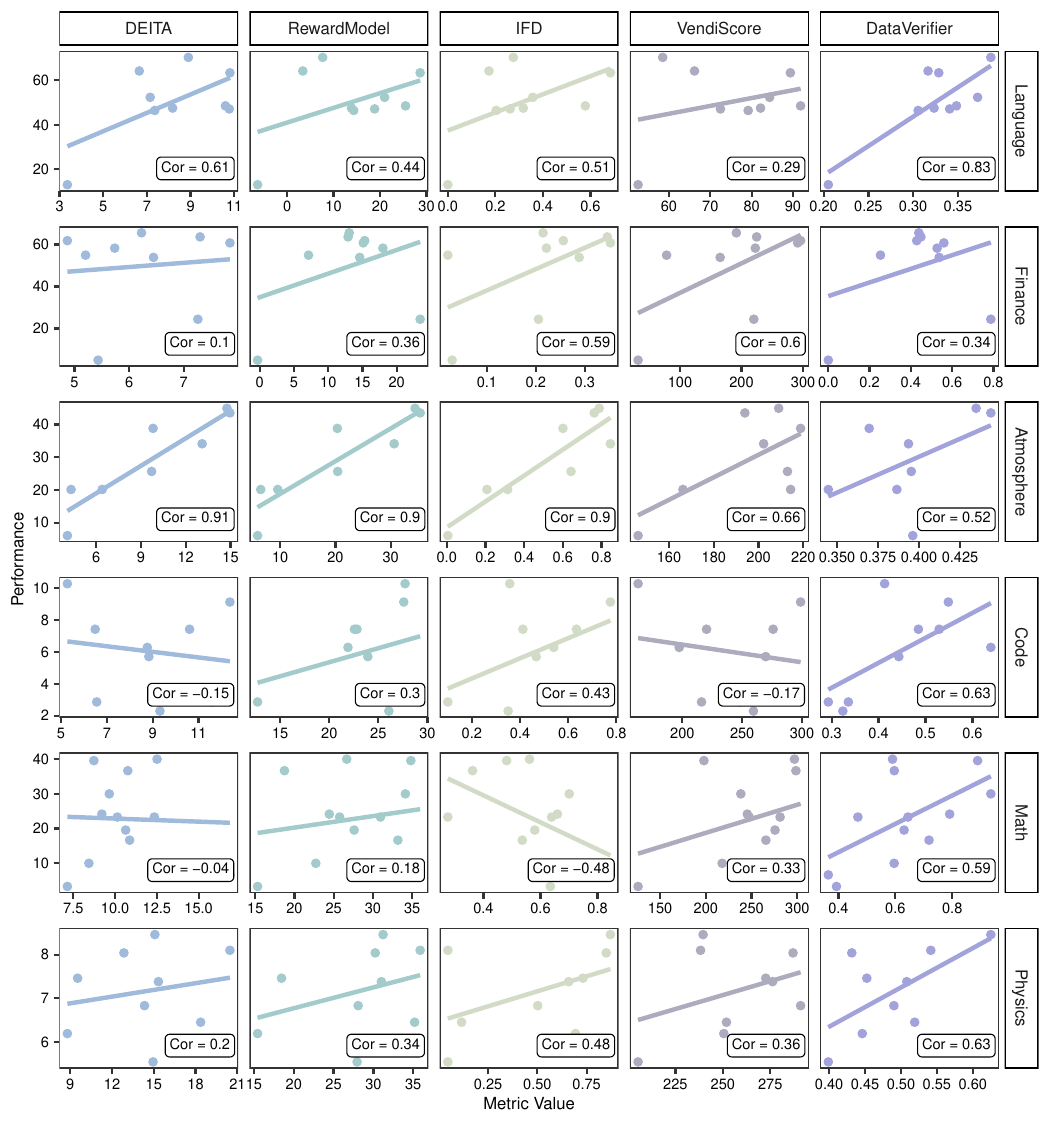}
    \caption{Complete results for correlation analysis.}
    \label{fig:appx-cor-all}
\end{figure*}

We provide the comprehensive correlation analysis results across all six evaluation tasks in Fig.~\ref{fig:appx-cor-all}. These results validate the robustness of our Data Verifier compared to baseline metrics across diverse domains.

\section{Computational Cost Analysis}
\label{appx:comp-resources}
For SFT, the average per-step time is 2.4 minutes with a batch size of 32 on 8 H200 GPUs. For RL, we allocate 8 H200 GPUs to the policy model and 2 H200 GPUs to the Data Verifier deployment. A single RL step, which processes 128 candidate recipes, takes an average of 20.2 minutes. Notably, the Data Verifier inference accounts for less than 2 minutes of this duration, demonstrating its high computational efficiency.

\end{document}